%% file: root.tex
\documentclass[letterpaper, 10 pt, conference]{ieeeconf}  

\IEEEoverridecommandlockouts                              
\usepackage{graphics} 
\usepackage{epsfig} 
\usepackage{mathptmx} 
\usepackage{times} 
\usepackage{amsmath} 
\usepackage{amssymb}  
\usepackage[table,usenames,dvipsnames]{xcolor}      
\usepackage[colorinlistoftodos]{todonotes}
\usepackage{subcaption}
\usepackage[noend]{algpseudocode}
\usepackage{algorithm}
\usepackage[normalem]{ulem}
\usepackage{array}
\usepackage{hyperref}
\makeatletter
\def\BState{\State\hskip-\ALG@thistlm}
\makeatother
\newcommand{\bx}{\mathbf{x}}
\newcommand{\hT}[1]{\mathbf{h}^{#1 T}}
\newcommand{\bH}{\mathbf{H}}

\begin{document}

\title{Unsupervised Deep Homography: A Fast and Robust Homography Estimation Model}

\author{Ty Nguyen$^*$, Steven W. Chen$^*$, Shreyas S. Shivakumar, Camillo J. Taylor, Vijay Kumar %
\thanks{The authors are with GRASP Lab, University of Pennsylvania, Philadelphia, PA 19104, USA, {\tt\small\{tynguyen, chenste, sshreyas, cjtaylor, kumar\}@seas.upenn.edu}.}
\thanks{$^*$: The authors have equal contributions}
}


\maketitle

\begin{abstract}
Homography estimation between multiple aerial images can provide relative pose estimation for collaborative autonomous exploration and monitoring. The usage on a robotic system requires a fast and robust homography estimation algorithm. In this study, we propose an unsupervised learning algorithm that trains a Deep Convolutional Neural Network to estimate planar homographies. We compare the proposed algorithm to traditional feature-based and direct methods, as well as a corresponding supervised learning algorithm. Our empirical results demonstrate that compared to traditional approaches, the unsupervised algorithm achieves faster inference speed, while maintaining comparable or better accuracy and robustness to illumination variation. In addition, our unsupervised method has superior adaptability and performance compared to the corresponding supervised deep learning method. Our image dataset and a Tensorflow implementation of our work are available at $https://github.com/tynguyen/unsupervisedDeepHomographyRAL2018$. 
\end{abstract}


\input{tex/introduction.tex}

\input{tex/problem.tex}

\input{tex/background.tex}

\input{tex/method.tex}

\input{tex/evaluation.tex}

\input{tex/conclusion.tex}
\bibliographystyle{IEEEtran}
\bibliography{ref/ref.bib}
\end{document}

%% file: tex/introduction.tex
\section{INTRODUCTION}
A homography is a mapping between two images of a planar surface from different perspectives. They play an essential role in robotics and computer vision applications such as image mosaicing~\cite{brown2003recognising}, monocular SLAM~\cite{shridhar2015monocular}, 3D camera pose reconstruction~\cite{zhang19963d} and virtual touring~\cite{pan2004easy, tang2007self}. For example, homographies are applicable in scenes viewed at a far distance by an arbitrary moving camera~\cite{capel2004image}, which are the situations encountered in UAV imagery. However, to work well in the aerial multi-robot setting, the homography estimation algorithm needs to be reliable and fast. 

The two traditional approaches for homography estimation are direct methods and feature-based methods~\cite{szeliski2006image}. Direct methods, such as the seminal Lucas-Kanade algorithm~\cite{Lucas:1981:IIR:1623264.1623280}, use pixel-to-pixel matching by shifting or warping the images relative to each other and comparing the pixel intensity values using an error metric such as the sum of squared differences (SSD). They initialize a guess for the homography parameters and use a search or optimization technique such as gradient descent to minimize the error function~\cite{baker2004lucas}. The robustness of direct methods can be improved by using different performance criterion such as the enhanced correlation coefficient (ECC)~\cite{evangelidis2008parametric}, integrating feature-based methods with direct methods~\cite{yan2014heask}, or by representing the images in the Fourier domain~\cite{lucey2013fourier}. In addition, the speed of direct methods can be increased by using efficient compositional image alignment schemes~\cite{munoz2015rationalizing}.

The second approach are feature-based methods. These methods first extract keypoints in each image using local invariant features (e.g. Scale Invariant Feature Transform (SIFT)~\cite{lowe2004distinctive}). They then establish a correspondence between the two sets of keypoints using feature matching, and use RANSAC~\cite{Fischler:1981:RSC:358669.358692} to find the best homography estimate. While these methods have better performance than direct methods, they can be inaccurate when they fail to detect sufficient keypoints, or produce incorrect keypoint correspondences due to illumination and large viewpoint differences between the images~\cite{wu2007improved}. In addition, these methods are significantly faster than direct methods but can still be slow due to the computation of the features, leading to the development of other feature types such as Oriented FAST and Rotated BRIEF (ORB)~\cite{rublee2011orb} which are more computationally efficient than SIFT, but have worse performance.

\begin{figure}[t]
  \centering
\begin{subfigure}{.9\linewidth}
  \centering
  \includegraphics[width=7cm,height=2.5cm]{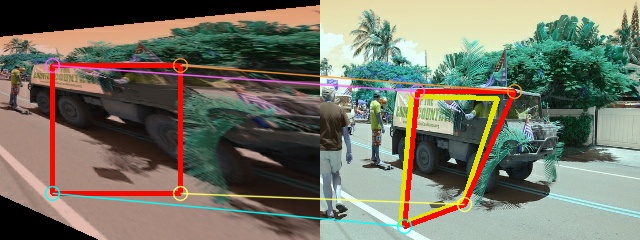}
\end{subfigure}  \\
    
\begin{subfigure}{.9\linewidth}
	\centering
    \includegraphics[width=7cm, height=2cm]{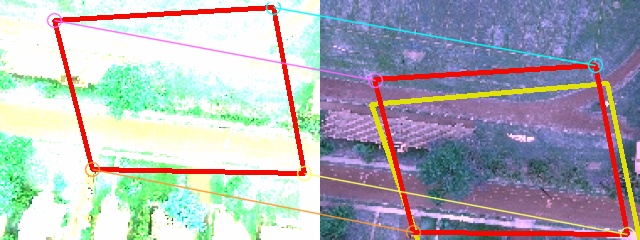}
\end{subfigure}   
\caption{Above: Synthetic data; Below: Real data; Homography estimation results from the unsupervised neural network. Red represents the ground truth correspondences, and yellow represents the estimated correspondences. These images depict an example of large levels of displacement and illumination shifts in which feature-based, direct and/or supervised learning methods fail.}
\label{fig:unsupervised_example}
\vspace{-2mm}
\end{figure}

Inspired by the success of data-driven Deep Convolutional Neural Networks (CNN) in computer vision, there has been an emergence of CNN approaches to estimating optical flow~\cite{weinzaepfel2013deepflow,ilg2016flownet,fischer2015flownet}, dense matching~\cite{revaud2016deepmatching,altwaijry2016learning}, depth estimation~\cite{eigen2014depth}, and homography estimation~\cite{detone2016}. Most of these works, including the most relevant work on homography estimation, treat the estimation problem as a supervised learning task. These supervised approaches use ground truth labels, and as a result are limited to synthetic datasets where the ground truth can be generated for free, or require costly labeling of real-world data sets. 

Our work develops an unsupervised, end-to-end, deep learning algorithm to estimate homographies. It improves upon these prior traditional and supervised learning methods by minimizing a pixel-wise intensity error metric that does not need ground truth data. Unlike the hand-crafted feature-based approaches, or the supervised approach that needs costly labels, our model is adaptive and can easily learn good features specific to different data sets. Furthermore, our framework has fast inference times since it is highly parallel. These adaptive and speed properties make our unsupervised networks especially suitable for real world robotic tasks, such as stitching UAV images. 

We demonstrate that our unsupervised homography estimation algorithm has comparable or better accuracy, and better inference speed, than feature-based, direct, and supervised deep learning methods on synthetic and real-world UAV data sets. In addition, we demonstrate that it can handle large displacements ($\sim 65\%$ image overlap) with large illumination variation. Fig.~\ref{fig:unsupervised_example} illustrates qualitative results on these data sets, where our unsupervised method is able to estimate the homography whereas the other approaches cannot.

Our unsupervised algorithm is a hybrid approach that combines the strengths of deep learning with the strengths of both traditional direct methods and feature-based methods. It is similar to feature-based methods because it also relies on features to compute the homography estimates, but it differs in that it learns the features rather than defining them. It is also similar to the direct methods because the error signal used to drive the network training is a pixel-wise error. However, rather than performing an online optimization process, it transfers the computation offline and "caches" the results through these learned features. Similar unsupervised deep learning approaches have been successful in computer vision tasks such as monocular depth and camera motion estimation~\cite{zhou2017unsupervised}, indicating that our framework can be scaled to tackle general nonlinear motions such as those encountered in optical flow.

%% file: tex/problem.tex
\section{Problem Formulation}
We assume that images are obtained by a perspective pin-hole camera and present 
points by homogeneous coordinates, so that a point $(u, v)^T$ is represented as $(u, v, 1)^T$ and a point $(x, y, z)^T$ is equivalent to the point $(x/z, y/z, 1)^T$. Suppose that $\mathbf{x} = (u, v, 1)^T$ and $\mathbf{x}' = (u', v', 1)^T$  are two points. A planar projective transformation or homography that maps $\mathbf{x} \leftrightarrow \mathbf{x}'$ is a linear transformation represented by a non-singular $3
\times 3$ matrix $\bH$ such that: 
\begin{align}
			\begin{bmatrix}
            u' \\ 
            v'   \\ 
            1  
			\end{bmatrix} 
          & = \begin{bmatrix}
            h_{11} & h_{12} & h_{13} \\ 
            h_{21} & h_{22} & h_{23} \\ 
            h_{31} & h_{32} & h_{33} 
			\end{bmatrix} 
            \begin{bmatrix}
            u \\ 
            v  \\ 
            1  
			\end{bmatrix} \hspace{.1in} \mbox{   Or  } \hspace{.1in} \mathbf{x}' = \mathbf{H} \mathbf{x}   
\label{eq:H_def}
\end{align}
Since $\bH$ can be multiplied by an arbitrary non-zero scale factor without altering the projective transformation, only the ratio of the matrix elements is significant, leaving $\bH$ eight independent ratios corresponding to eight degrees of freedom. This mapping equation can also represented by two equations: 
\begin{align}
 u' = \frac{h_{11}u + h_{12}v + h_{13}}{h_{31}u + h_{32}v + h_{33}} ;  
 v' = \frac{h_{21}u + h_{22}v + h_{23}}{h_{31}u + h_{32}v + h_{33}} 
 \label{eq:close_H}
\end{align}

The problem of finding the homography  induced by two images $I^A$ and $I^B$ is to find a homography $\bH^{AB}$ such that Eqn.~\eqref{eq:H_def} holds for all points in the overlapping of the two images.

%% file: tex/background.tex
\section{Supervised Deep Homography Model}
\begin{figure*}[t]
\centering
\vspace{2mm}
\includegraphics[width=0.96\textwidth]{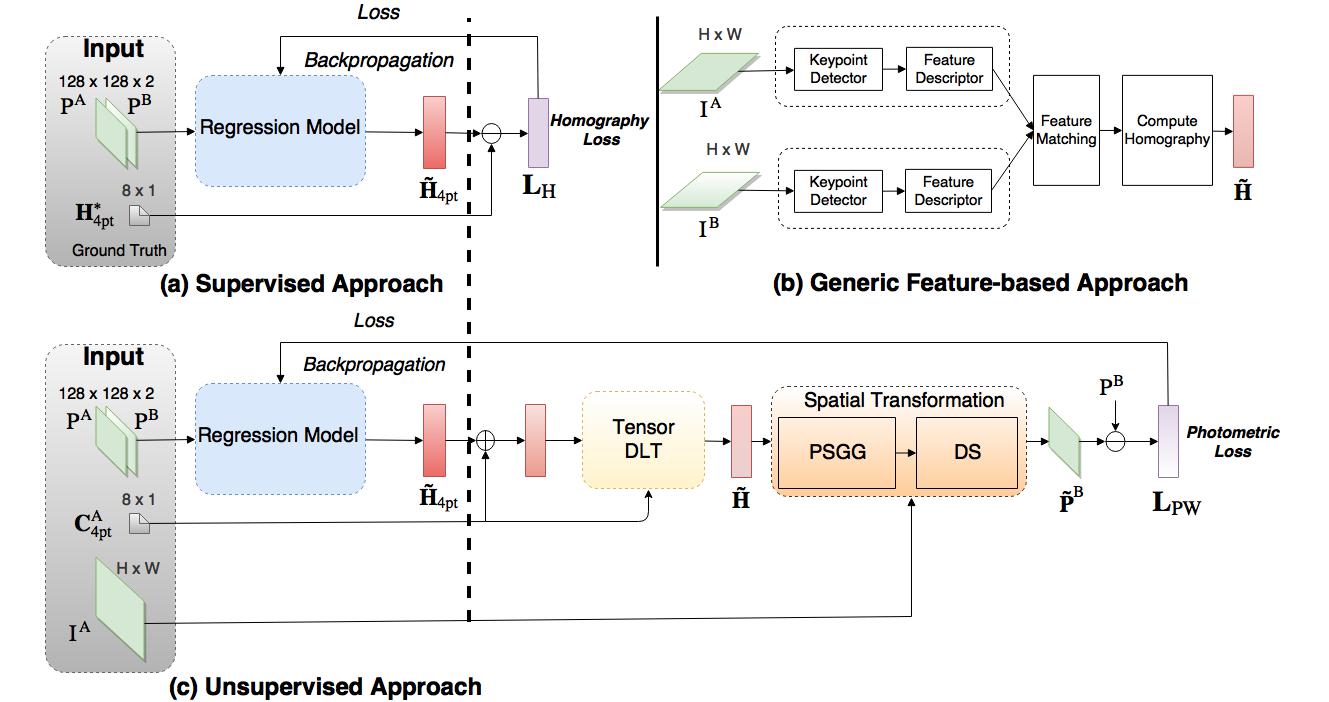}
\caption{Overview of homography estimation methods; (a) Benchmark supervised deep learning approach; (b) Feature-based methods; and (c) Our unsupervised method. DLT: direct linear transform; 
PSGG: parameterized sampling grid generator; DS: differentiable sampling. 
\label{fig:methods}
}
\vspace{-2mm}
\end{figure*}

The deep learning approach most similar to our work is the Deep Image Homography Estimation~\cite{detone2016}. In this work, DeTone et al. use supervised learning to train a deep neural network on a synthetic data set. They use the 4-point homography parameterization $\mathbf{H}_{4pt}$~\cite{baker2006parameterizing} rather than the conventional $3 \times 3$ parameterization $\mathbf{H}$. Suppose that $\mathbf{u}^A_k = (u^A_k, v_k^A,1)^T$ and $\mathbf{u}_k^B = (u_k^B, v_k^B,1)^T$ for $k = 1, 2, 3, 4$ are 4 fixed points in image $I^A$  and $I^B$ respectively, such that $\mathbf{u}_2^k = \mathbf{H} \mathbf{u}_1^k.$ Let $\Delta u_k = u_k^B -   u_k^A$, $\Delta v_k = v_k^B -   v_k^A$. Then $\mathbf{H}_{4pt}$ is the $4\times 2$ matrix of points $(\Delta u_k, \Delta v_k)$. Both parameterizations are equivalent since there is a one-to-one correspondence between them. 

In a deep learning framework though, this parameterization is more suitable than the $3 \times 3$ parameterization $\mathbf{H}$ because $\mathbf{H}$ mixes the rotation, translation, scale, and shear components of the homography transformation. The rotation and shear components tend to have a much smaller magnitude than the translation component, and as a result although an error in their values can greatly impact $\mathbf{H}$, it will have a small effect on the $L2$ loss function of the elements of $\mathbf{H}$, which is detrimental for training the neural network. In addition, the high variance in the magnitude of the elements of the $3 \times 3$ homography matrix makes it difficult to enforce $\mathbf{H}$ to be non-singular. The $4$-point parameterization does not suffer from these problems.

The network architecture is based on VGGNet~\cite{simonyan2014very}, and is depicted in Fig.~\ref{fig:methods}(a). The network input is a batch of image patch pairs. The patch pairs are generated by taking a full-sized image, cropping a square patch $P^A$ at a random position $p$, perturbing the four corners of by a random value within $[-\rho, \rho]$ to generate a homography $\mathbf{H}^{AB}$, applying $(\mathbf{H}^{AB})^{-1}$ to the full-sized image, and then cropping a square patch $P^B$ of the same size and at the same location as the patch $P^A$ from the warped image. These image patches are used to avoid strange border effects near the edges during the synthetic data generation process, and to standardize the network input size. The applied homography $\mathbf{H}^{AB}$ is saved in the $4$ point parameterization format, $\mathbf{H}^{*}_{4pt}$. The network outputs a  $4$ point parameterization estimate $\mathbf{\tilde{H}}_{4pt}$.

The error signal used for gradient backpropagation is the Euclidean $L2$ norm, denoted as $\mathbf{L}_H$, of the estimated $4$-point homography $ \mathbf{\tilde{H}}_{4pt}$ versus the ground truth $\mathbf{H}^{*}_{4pt}$:
\begin{equation}
\mathbf{L}_H = \frac{1}{2} || \mathbf{\tilde{H}}_{4pt}  - \mathbf{H}^{*}_{4pt} || ^2_2 
\label{eq:H_loss}
\end{equation}

%% file: tex/method.tex
\section{Unsupervised Deep Homography Model}
\label{sec:method}
While the supervised deep learning method has promising results, it is limited in real world applications since it requires ground truth labels. Drawing inspiration from traditional direct methods for homography estimation, we can define an analogous loss function. Given an image pair $I^{A}(\mathbf{x})$ and $I^{B}(\mathbf{x})$ with discrete pixel locations represented by homogeneous coordinates $\{\mathbf{x}_{i} = (x_{i}, y_{i}, 1)^{T}\}$, we want our network to output $\mathbf{\tilde{H}}_{4pt}$ that minimizes the average $L1$ pixel-wise photometric loss
\begin{equation}
\mathbf{L}_{PW} =  \frac{1}{|\mathbf{x}_{i}|}\sum_{\mathbf{x}_{i}} | I^{A}(\mathcal{H}(\mathbf{x}_{i})) - I^{B}(\mathbf{\mathbf{x}_{i}})|
\label{eq:pwrmse_loss}
\end{equation}
where $\mathbf{\tilde{H}}_{4pt}$ defines the homography transformation $\mathcal{H}(\mathbf{x}_{i})$. We chose the $L1$ error versus the $L2$ error because previous work has observed that it is more suitable for image alignment problems~\cite{zhao2015l2}, and empirically we found the network to be easier to train with the $L1$ error. This loss function is unsupervised since there is no ground truth label. Similar to the supervised case, we choose the $4$-point parameterization which is more suitable than the $3\times 3$ parameterization.

In order to compare our unsupervised deep learning algorithm with the supervised algorithm, we use the same VGGNet architecture to output the $\mathbf{\tilde{H}}_{4pt}$. Fig.~\ref{fig:methods}(c) depicts our unsupervised learning model. The regression module represents the VGGNet architecture and is shared by both the supervised and unsupervised methods. Although we do not investigate other possible architectures, different regression models such as SqueezeNet~\cite{iandola2016squeezenet} may yield better performance due to advantages in size and computation requirements. The second half of Fig.~\ref{fig:methods}(c) represents the main contribution of this work, which consists of the differentiable layers that allow the network to be successfully trained with the loss function~\eqref{eq:pwrmse_loss}.

Using the pixel-wise photometric loss function yields additional training challenges. First, every operation, including the warping operation $\mathcal{H}(\mathbf{x}_{i})$, must remain differentiable to allow the network to be trained via backpropagation. Second, since the error signal depends on differences in image intensity values rather than the differences in the homography parameters, training the deep network is not necessarily as easy or stable. Another implication of using a pixel-wise photometric loss function is the implied assumption that lighting and contrast between the input images remains consistent. In traditional direct methods such as ECC, this appearance variation problem is addressed by modifying the loss function or preprocessing the images. In our unsupervised algorithm, we standardize our images by the mean and variance of the intensities of all pixels in our training dataset, perform data augmentation by injecting random illumination shifts, and use the standard $L1$ photometric loss. We found that even without modifying the loss function, our deep neural network is still able to learn to be invariant to illumination changes.
\subsection{Model Inputs}
The input to our model consists of three parts. The first part is a 2-channel image of size $128 \times 128 \times 2$ which is the stack of $P^A$ and $P^B$ - two patches cropped from the two images $I^A$ and $I^B$. The second part is the four corners in $I^A$, denoted as $\mathbf{C}^A_{4pt}$. Image $I^A$ is also part of the input as it is necessary for warping.
 
\input{tex/mapping.tex}

\input{tex/warping.tex}

%% file: tex/mapping.tex
\subsection{Tensor Direct Linear Transform}
\label{sec:mapping}
We develop a Tensor Direct Linear Transform (Tensor DLT) layer to compute a differentiable mapping from the 4-point parameterization $\mathbf{\tilde{H}_{4pt}}$ to $\mathbf{\tilde{H}}$, the $3 \times 3$ parameterization of homography. This layer essentially applies the DLT algorithm~\cite{Hartley2004} to tensors, while remaining differentiable to allow backpropagation during training. As shown in Fig. ~\ref{fig:methods}(c), the input to this layer are the corresponding corners in the image pairs $\mathbf{C}^A_{4pt}$ and $\mathbf{\tilde{C}}^{B}_{4pt}$, and the output is the estimate of the $3\times 3$ homography parameterization $\mathbf{\tilde{H}}$. 

The DLT algorithm is used
to solve for the homography matrix $\mathbf{H}$ given a set of four point correspondences~\cite{Hartley2004}. Let $\mathbf{H}$ be the homography induced by a set of four 2D to 2D correspondences, $\mathbf{x}_i \leftrightarrow \mathbf{x}'_i$. According to the definition of a homography given in Eqn.~\eqref{eq:H_def}, $\bx'_i \sim \bH \bx_i$. This relation can also be expressed as $\bx'_i \times \mathbf{H} \bx_i = 0$. 

Let $\hT{j}$ be the $j$-th row of $\bH$, then: 
\begin{equation}
\bH \bx_i = \begin{bmatrix}
        \hT{1} \bx_i \\ 
                \hT{2} \bx_i \\ 
                \hT{3} \bx_i 
      \end{bmatrix}
         = \begin{bmatrix}
        \bx_i^T \mathbf{h}^1  \\ 
                \bx_i^T \mathbf{h}^2  \\ 
                \bx_i^T \mathbf{h}^3  
      \end{bmatrix}
\end{equation}
where $\mathbf{h}^j$ is the column vector representation of $\hT{j}$. 

Let $\bx'_i = (u'_i, v'_i, 1)^T$, then:
\begin{equation}
\bx'_i \times \bH \bx_i = \begin{bmatrix}
         v'_i \bx_i^T \mathbf{h}^3  -  \bx_i^T \mathbf{h}^2  \\ 
               \bx_i^T \mathbf{h}^1  -        u'_i\bx_i^T \mathbf{h}^3  \\ 
               u'_i \bx_i^T \mathbf{h}^2  -   v'_i\bx_i^T \mathbf{h}^1  
                    \end{bmatrix} = 0
\end{equation}

This equation can be rewritten as: 
\begin{equation}
               \begin{bmatrix}
         0_{3\times 1}^T     &  -\bx_i^T             &  v'_i \bx_i^T \\ 
               \bx_i^T             &    0_{3\times 1}^T    & -u'_i\bx_i^T \\        
               - v'_i\bx_i^T       &  u'_i \bx_i^T         &    0_{3\times 1}^T 
           \end{bmatrix}
               \begin{bmatrix}
         \mathbf{h}^1  \\ 
                 \mathbf{h}^2  \\ 
                 \mathbf{h}^3  
      \end{bmatrix}  = 0. 
\label{eq:3_9}
\end{equation}
which has the form $\mathbf{A}^{(3)}_i \mathbf{h} = \mathbf{0}$ for each $i=1,2,3,4$ correspondence pair, where $\mathbf{A}^{(3)}_i$ is a $3 \times 9$ matrix, and $\mathbf{h}$ is a vector with $9$ elements consisting of the entries of $\bH$. Since the last row in $\mathbf{A}^{(3)}_{i}$ is dependent on the other rows, we are left with two linear equations $\mathbf{A}_i \mathbf{h} = \mathbf{0}$ where $\mathbf{A}_i$ is the first 2 rows of $\mathbf{A}^{(3)}_i$.

Given a set of 4 correspondences, we can create a system of equations to solve for $\mathbf{h}$ and thus $\bH$. For each $i$, we can stack $\mathbf{A}_i$ to form $\mathbf{A} \mathbf{h} = \mathbf{0}$. Solving for $\mathbf{h}$ results in finding a vector in the null space of $\mathbf{A}$. One popular approach is singular value decomposition (SVD)~\cite{golub1970singular}, which is a differentiable operation. However, taking the gradients in SVD has high time complexity and has practical implementation issues~\cite{papadopoulo2000estimating}. An alternative solution to this problem is to make the assumption that the last element of $\mathbf{h}^{3}$, which is $\bH_{33}$ is equal to 1~\cite{hartley2003multiple}.  

With this assumption and the fact that $\mathbf{x}_{i}=(u_{i}, v_{i}, 1)$, we can rewrite Eqn.~\eqref{eq:3_9} in the form $\mathbf{\hat{A}}_i \mathbf{\hat{h}} = \mathbf{\hat{b}}_{i}$ for each $i=1,2,3,4$ correspondence points where $\mathbf{\hat{A}}_{i}$ is the $2 \times 8$ matrix representing the first $8$ columns of $\mathbf{A}_{i}$,
\begin{equation}
               \mathbf{\hat{A}}_{i} = \begin{bmatrix}
         0 & 0 & 0  &  -u_i & -v_i & -1 & v'_i u_i & v'_i v_i\\ 
               u_i & v_i & 1 & 0 & 0 & 0 & -u'_i u_i & -u'_i v_i\\        
           \end{bmatrix}, \nonumber 
\label{eq:gen_Ai}
\end{equation}  
$\mathbf{\hat{b}}_{i}$ is a vector with 2 elements representing the last column of $\mathbf{A}_{i}$ subtracted from both sides of the equation, 
\begin{equation}
   \mathbf{\hat{b}}_{i} = [-v'_i, u'_i]^T
   \nonumber,
 \label{eq:gen_bi}
\end{equation}  
and $\mathbf{\hat{h}}$ is a vector consisting of the first 8 elements of $\mathbf{h}$ (with $\mathbf{H}_{33}$ omitted).

By stacking these equations, we get:
\begin{equation}
\mathbf{\hat{A}} \mathbf{\hat{h}} = \mathbf{\hat{b}},
\label{eq:ah_b}
\end{equation}

Eqn.~\eqref{eq:ah_b} has a desirable form because $\mathbf{\hat{h}}$, and thus $\mathbf{H}$, can be solved for using $\mathbf{\hat{A}}^{+}$, the pseudo-inverse of $\mathbf{\hat{A}}$.  This operation is simple and differentiable with respect to the coordinates of $\bx_i$ and $\bx'_i$. In addition, the gradients are easier to calculate than for SVD.

This approach may still fail if the correspondence points are collinear: if three of the correspondence points are on the same line, then solving for $\mathbf{H}$ is undetermined. We alleviate this problem by first making the initial guess of $\mathbf{H}_{4pt}$ to be zero, implying that $\mathbf{\tilde{C}}^{B}_{4pt} \sim \mathbf{C}^A_{4pt}$. We then set a small learning rate such that after each training iteration, $\mathbf{\tilde{C}}^{B}_{4pt}$ does not move too far away from  $\mathbf{C}^A_{4pt}$.

%% file: tex/warping.tex
  


\subsection{Spatial Transformation Layer}
The next layer applies the $3\times 3$ homography estimate $\mathbf{\tilde{H}}$ output by the Tensor DLT to the pixel coordinates $\mathbf{x}_{i}$ of image $I^{A}$ in order to get warped coordinates $\mathcal{H}(\mathbf{x}_{i})$. These warped coordinates are necessary in computing the photometric loss function in Eqn.~\eqref{eq:pwrmse_loss} that will train our neural network. In addition to warping the coordinates, this layer must also be differentiable so that the error gradients can flow through via backpropagation. We thus extend the Spatial Transformer Layer introduced in~\cite{jaderberg2015spatial} by applying it to homography transformations.

This layer performs an inverse warping in order to avoid holes in the warped image. This process consists of 3 steps: (1) Normalized inverse computation $\mathbf{\tilde{H}}_{inv}$ of the homography estimate; (2) Parameterized Sampling Grid Generator (PSGG); and (3) Differentiable Sampling (DS). 

The first step, computing a normalized inverse, involves normalizing the height and width coordinates of images $I^{A}$ and $I^{B}$ into a range such that $-1 \leq u_i, v_i \leq 1$ and $-1 \leq u'_i, v'_i \leq 1$. Thus given a $3 \times 3$ homography estimate $\mathbf{\tilde{H}}$, the inverse $\mathbf{\tilde{H}}_{inv}$ used for warping is computed as follows:
\begin{align*}
\mathbf{\tilde{H}}_{inv} =&\ M^{-1} \mathbf{\tilde{H}}^{-1}  M \\ 
\mbox{where } M = &\ \begin{bmatrix}
               W'/2 & 0 & W'/2 \\
               0     & H'/2 & H'/2 \\ 
               0   &     0      &   1
             \end{bmatrix}
\end{align*}
with $W'$ and $H'$ are the width and height of the $I_{B}$. 

The second step (PSGG) creates a grid $G = \{G_i\}$ of the same size as the second image $I^{B}$. Each grid element $G_i = (u'_i, v'_i)$ corresponds to pixels of the second image $I^{B}$. Applying the inverse homography $\mathbf{\tilde{H}}_{inv}$ to these grid coordinates provides a grid of pixels in the first image $I^{A}$. 
\begin{equation}
            \begin{bmatrix}
            u_i \\ 
            v_i  \\
            1 
			\end{bmatrix} = \mathcal{H}_{inv} (G_i) = \mathbf{\tilde{H}}_{inv}
            \begin{bmatrix}
            u'_i \\ 
            v'_i   \\ 
            1  
			\end{bmatrix} 
\end{equation}

Based on the sampling points $\mathcal{H}_{inv} (G_i)$ computed from PSGG, the last step (DS) produces a sampled warped image $V$ of size $H' \times W'$ with $C$ channels, where $V(\mathbf{x}_{i}) = I^{A}(\mathcal{H}(\mathbf{x}_{i}))$. 

  


The sampling kernel $k(\cdot)$ is applied to the grid $\mathcal{H}_{inv} (G_i)$ and the resulting image $V$ is defined as
\begin{align}
V^C_i = \sum_n^H \sum_m^W I^c_{nm} k (u_i - m; \Phi_u) k(v_i-n; \Phi_v), \nonumber \\ \forall i \in [1...H'W'] \hspace{.1in}, \forall c \in [1..C]
\label{eq:interpolation}
\end{align}
where $H, W$ are the height and width of the input image $I^{A}$, $\Phi_u$ and $\Phi_v$ are the parameters of $k(\cdot)$ defining the image interpolation. $I^c_{nm}$ is the value at location $(n,m)$ in channel $c$ of the input image, and $V^c_i$ is the value of the output pixel at location $(u_i,v_i)$ in channel $c$. Here, we use bilinear interpolation such that the Eqn.~\eqref{eq:interpolation} becomes 
\begin{align}
V^C_i = \sum_n^H \sum_m^W I^c_{nm}  \max(0, 1 - |u_i - m|) \max (0, 1 - |v_i-n|)
\label{eq:bilinear}
\end{align}
To allow backpropagation of the loss function, gradients with respect to $I$ and $G$ for bilinear interpolation are defined as 
\begin{align}
	\frac{\partial V^c_i}{\partial I^c_{nm}} & = \sum_n^H \sum_m^W   \max(0, 1 - |u_i - m|) \max (0, 1 - |v_i-n|) \\ 
    \frac{\partial V^c_i}{\partial u^i} & = \sum_n^H \sum_m^W I^c_{nm} \max (0, 1 - |v_i-n|)         \left\{
                \begin{array}{ll}
                  0   \mbox{ if } |m - u_i| \geq 1   \\
                  1    \mbox{ if } m \geq u_i \\
                  - 1   \mbox{ if } m < u_i
                \end{array}
              \right. \\ 
   \frac{\partial V^c_i}{\partial v^i} & = \sum_n^H \sum_m^W I^c_{nm} \max (0, 1 - |u_i-m|)         \left\{
                \begin{array}{ll}
                  0   \mbox{ if } |n - v_i| \geq 1   \\
                  1    \mbox{ if } n \geq v_i \\
                  - 1   \mbox{ if } n < v_i
                \end{array}
              \right.
\end{align}
This allows backpropagation of the loss gradients using the chain rule because $\frac{\partial u_i}{\partial h_{jk}}$ and $\frac{\partial v_i}{\partial h_{jk}}$ can be easily derived from Eqn.~\ref{eq:close_H}.

%% file: tex/evaluation.tex
\section{Evaluation Results}
\begin{figure*}[t]
  \centering
    \begin{minipage}{.3\textwidth}
      \vspace{-2mm}
      \includegraphics[width=\linewidth]{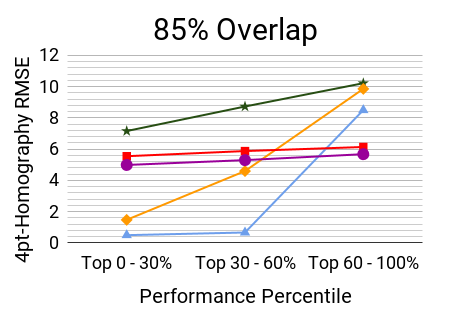}
  
    \end{minipage}
    \begin{minipage}{.3\textwidth}
       \vspace{-2mm}
       \includegraphics[width=\linewidth]{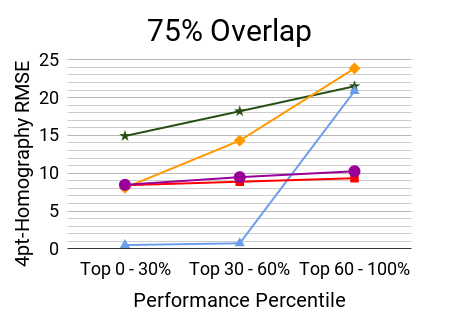}
    \end{minipage} 
    \begin{minipage}{.36\textwidth}
       \vspace{-2mm}
       \includegraphics[width=\linewidth]{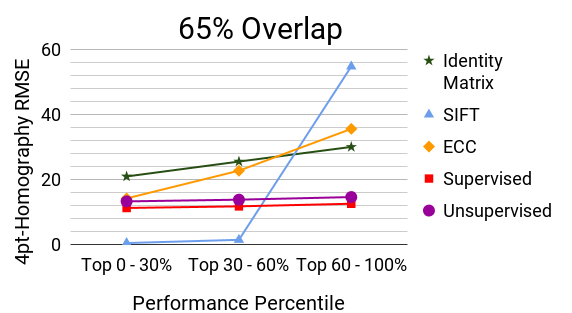}
    \end{minipage}  
  \caption{Synthetic 4pt-Homography RMSE (lower is better). Unsupervised has comparable performance with the supervised method and performs better than the other approaches especially when the displacement is large.}
\label{fig:4pt_RMSE_syn_variant_img}
\end{figure*} 

\begin{figure}[h]
\centering
\includegraphics[width=0.95\linewidth]{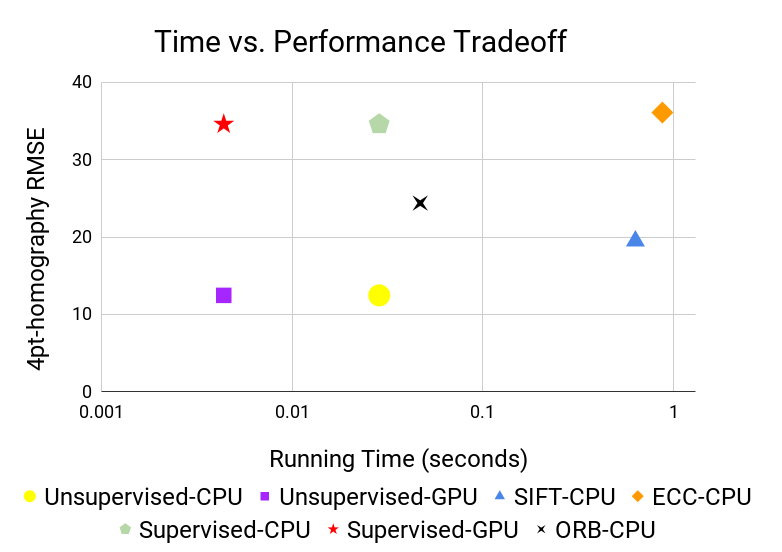}
\caption{Speed Versus Performance Tradeoff. Lower left is better. Suffixes GPU and CPU reflect the computational resource. All the feature-based methods are run on the CPU. The unsupervised network run on the GPU dominates all the other methods by having both the highest throughput and best performance.} 
\label{fig:speed_vs_performance}
\vspace{-4mm}
\end{figure}

\begin{figure*}[t!]
  \centering
    \begin{minipage}{.45\textwidth}
      \subcaption{Unsupervised $(Success, RMSE = 15.6)$}
      \vspace{-2mm}
      \includegraphics[width=\linewidth]{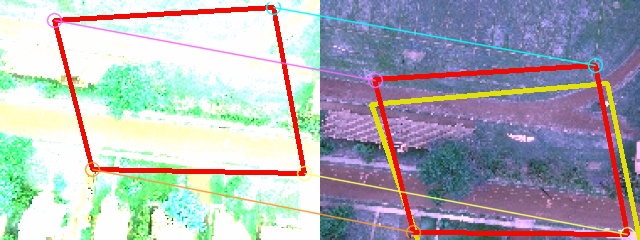}
  
    \end{minipage}
    \begin{minipage}{.45\textwidth}
       \subcaption{Unsupervised $(Success, RMSE = 4.50)$}
       \vspace{-2mm}
       \includegraphics[width=\linewidth]{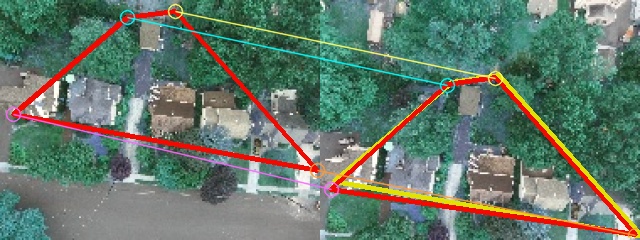}
      
    \end{minipage} \\  
    \begin{minipage}{.45\textwidth}
      \vspace{.5mm}
      \subcaption{SIFT $(Fail, RMSE = 105.2)$}
      \vspace{-2mm}
      \includegraphics[width=\linewidth]{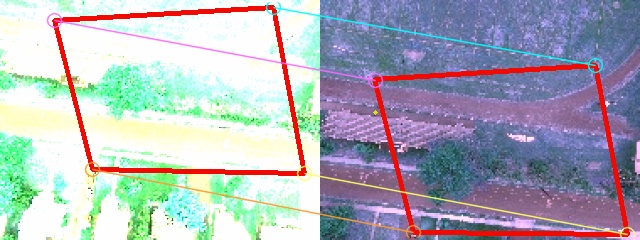}
    \end{minipage} 
    \begin{minipage}{.45\textwidth}
       \vspace{.5mm}
       \subcaption{SIFT $(Success, RMSE = 6.06)$}
       \vspace{-2mm}
       \includegraphics[width=\linewidth]{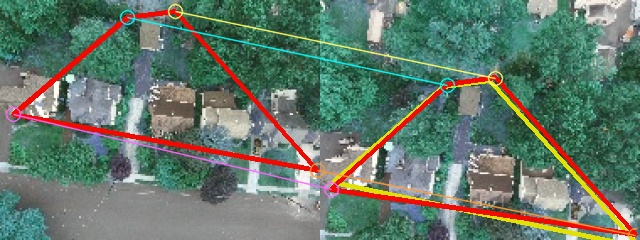}
    \end{minipage}  \\
  
    \begin{minipage}{.45\textwidth}
      \vspace{.5mm}
      \subcaption{ECC$(Success,  RMSE = 66.4)$}
      \vspace{-2mm}
      \includegraphics[width=\linewidth]{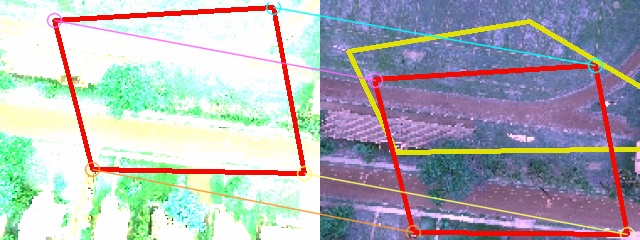}
    \end{minipage}
    \begin{minipage}{.45\textwidth}
      \vspace{.5mm}
      \subcaption{ECC$(Success, RMSE = 48.10)$}
      \vspace{-2mm}
       \includegraphics[width=\linewidth]{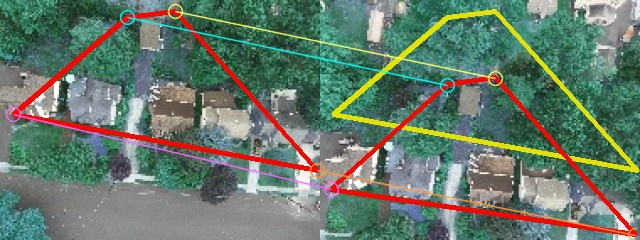}
    \end{minipage} 
  \caption{Qualitative visualization of estimation methods on aerial dataset. Left: hard case, right: moderate case. ECC performs better than SIFT in the case of small displacement, but performs worse than SIFT in case of large displacement. Unsupervised network outperforms both SIFT and ECC approaches. Supervised network is omitted due to limited space and its poor performance on this dataset. }\label{fig:compare_homography_real_images}
\vspace{-2mm}
\end{figure*}

The intended use case for our algorithm is in estimating homographies for aerial multi-robot systems applications such as image mosaicing and collision avoidance. As a result, we demonstrate our unsupervised algorithm's accuracy, inference speed, and robustness to illumination variation relative to SIFT, ORB, ECC and the supervised deep learning method. We evaluate these methods on a synthetic dataset similar to the dataset used in~\cite{detone2016}, and on a real-world aerial image dataset. Since ORB's performance is inferior to that of SIFT, we only report ORB's performance in Fig.~\ref{fig:speed_vs_performance} and omit it in the remaining figures. 

Both the supervised and unsupervised approaches use the VGGNet architecture to generate homography estimates $\mathbf{\tilde{H}_{4pt}}$. The deep learning approaches are implemented in Tensorflow~\cite{abadi2016tensorflow} using stochastic gradient descent with a batch size of 128, and an Adam Optimizer~\cite{kingma2014adam} with, $\beta_1 = 0.9$, $\beta_2 = 0.999$ and $\epsilon = 10^{-8}$. We empirically chose the initial learning rate for the supervised algorithm and unsupervised algorithm to be $0.0005$ and $0.0001$ respectively.

The ECC direct method is a standard Python OpenCV implementation while the feature-based approaches are Python OpenCV implementations of SIFT RANSAC and ORB RANSAC. We found that in our synthetic dataset, using all detected features gives better performance, while in our aerial dataset, choosing the 50 best features is superior. These feature pairs are then used to calculate the homography using RANSAC with a threshold of $5$ pixels. For the ECC method, we use identity matrix as the initialization and set 1000 as the maximum number of iterations. 

\subsection{Synthetic Data Results}
This section analyzes the performance profile of the Unsupervised, Supervised, SIFT, and ECC methods on our synthetic dataset. We want to test how well our approach performs under illumination variation and large image displacement.

To account for illumination variation, we globally standardize our images based on the mean and variance of pixel intensities of all images in our training dataset. We additionally inject random color, brightness and gamma shifts during the training. We do not utilize any further preprocessing and use the $L1$ photometric loss function. To highlight the effect of displacement amount on each method, we break down the accuracy performance in terms of: $85\%$ image overlap (small displacement), $75\%$ image overlap (moderate displacement), and $65\%$ image overlap (large displacement). We follow the synthetic data generation process on the MS-COCO dataset used in~\cite{detone2016}. The amount of image overlap is controlled by the point perturbation parameter $\rho$. The evaluation metric is the $4$pt-Homography RMSE from Eqn.~\eqref{eq:H_loss} comparing the estimated homography to the ground truth homography. 

We train the deep networks from scratch for $300,000$ iterations over $\sim 30$ hours, using two GPUs. This long training procedure only needs to be performed once, as the resulting model can be used as an initial pre-trained model for other data sets. We observed that the supervised model started overfitting after $150,000$ iterations so stopped training early. SIFT, ORB and ECC estimated homographies using the full images, while the deep learning methods are only given access to the small patches ($\sim 21 \%$ pixels). This disadvantages our methods, and would result in better performance for the traditional methods, at the expense of slower running times.

\begin{figure}
\centering
\includegraphics[width=0.95\linewidth]{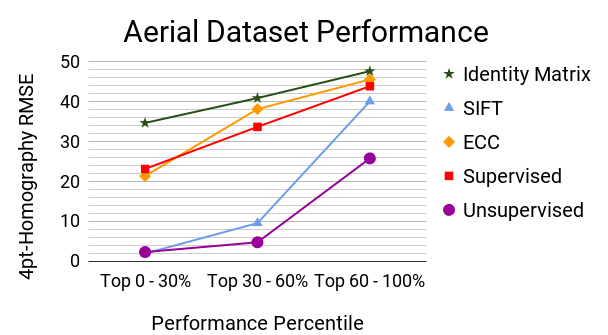}
\caption{4pt-homography RMSE on aerial images (lower is better). Unsupervised outperforms other approaches significantly.}
\label{fig:4pt_RMSE_real_img}
\vspace{-4mm}
\end{figure}

Fig.~\ref{fig:4pt_RMSE_syn_variant_img} displays the results of each method broken down by overlap and performance percentile. We break down the results by performance percentile to illustrate the various performance profiles of each method. Specifically, SIFT tends to do very well $60 \%$ of the time, but in the worst $40 \%$ of the time it performs very poorly, sometimes completely failing to detect enough features to estimate a homography. On the other hand, the deep learning methods tends to have much more consistent performance, which can be more desireable in practical applications such as using homographies for collision avoidance for aerial multi-robot systems. Both the learning methods and the feature-based methods outperform the direct method (ECC). 

Interestingly, whereas direct method ECC has problems with illumination variation and large displacement, our unsupervised method is able to handle these scenarios even though it uses photometric loss functions. One potential hypothesis is that our method can be viewed as a hybrid between direct methods and feature based methods. The large receptive field of neural networks may allow it to handle large image displacement better than a direct method. In addition, whereas image gradients are used to update homography parameters in direct methods, with neural networks, these gradients are used to update network parameters which correspond to improving learned features. Finally, direct methods are an online optimization process that use gradients from a single pair of images, whereas training a deep network is an offline optimization process that averages gradients across multiple images. Injecting noise into this training process can further improve robustness to different appearance variations. Understanding the relationship between the neural network and photometric loss functions is an important direction for future work.

\subsection{Aerial Dataset Results}
This section analyzes the performance profile of each method on a representative dataset of aerial imagery captured by a UAV. In addition to accuracy performance, an equally important consideration for real world application is inference speed. As a result, we also discuss the performance to speed tradeoffs of each method.

Our aerial dataset contains $350$ image pairs resized to $240 \times 320$, captured by a DJI Phantom $3$ Pro platform in Yardley, Pensylvania, USA in $2017$. We divided it into $300$ train and $50$ test samples. We did not label the train set, but for evaluation purposes, we manually labeled the ground truth by picking 4 pairs of correspondences for each test sample. We also randomly inject illumination noise in both the training and testing sets. The evaluation metrics are the same for the synthetic data. To reduce training time, we finetune the neural networks on the aerial image data. Our unsupervised algorithm can directly use the aerial dataset image pairs. However, since we do not have ground truth homography labels, we have to perform a similar synthetic data generation process as in the synthetic dataset in order to finetune the supervised neural network. We fine tune both models over $150,000$ iterations for roughly $15$ hours with data augmentation.

Fig.~\ref{fig:4pt_RMSE_real_img} displays the performance profile for the Unsupervised, Supervised, SIFT, and ECC methods. Fig~\ref{fig:speed_vs_performance} displays the speed and performance tradeoff for these methods, and additionally the featured based method ORB. The feature-based methods are tested on a 16-core Intel Xeon CPU, and the deep learning methods are tested on the same CPU and an NVIDIA Titan X GPU. The closer to the lower left hand corner, the better the performance and faster the runtime.

Both Figs.~\ref{fig:4pt_RMSE_real_img} and~\ref{fig:speed_vs_performance} demonstrate that our unsupervised algorithm has the best performance of all methods. In addition, Fig.~\ref{fig:speed_vs_performance} also shows that our unsupervised method on the GPU has both the best performance and the fastest inference times. SIFT has the second best performance after our unsupervised algorithm, but has a much slower runtime (approximately $200$ times slower). ORB has a faster runtime than SIFT, but at the expense of poorer performance. The ECC direct method approach has the worst performance and runtime of all the methods. A qualitative example where both SIFT and ECC fail to deliver a good result while our method succeeds is illustrated in~Fig.~\ref{fig:compare_homography_real_images}.

One of the most interesting results is that while the supervised and unsupervised approaches performed comparably on the synthetic data, the supervised approach had drastically poorer performance on the aerial image dataset. This shift is due to the fact that ground truth labels are not available for our aerial dataset. The generalization gap from synthetic (train) to real (test) data is an important problem in machine learning. The best practical approach is to additionally fine-tune the model on the new distribution of data. In a robotic field experiment, this can be achieved by flying the UAV to collect a few sample images and fine-tuning on those images. However, this fine-tuning is only possible with our unsupervised algorithm. Our aerial dataset results highlight the fact that even though synthetic data can be generated from real images, a pair of synthetic images is still very different from a pair of real images. These results demonstrate that the independence of our unsupervised algorithm from expensive ground truth labels has large practical implications for real-world performance.

%% file: tex/conclusion.tex
\section{Conclusions}
We have introduced an unsupervised algorithm that trains a deep neural network to estimate planar homographies. Our approach outperforms the corresponding supervised network on both synthetic and real-world datasets, demonstrating the superiority of unsupervised learning in image warping problems. Our approach achieves faster inference speed, while maintaining comparable or better accuracy than feature-based and direct methods. We demonstrate that the unsupervised approach is able to handle large displacements and large illumination variations that are typically challenging for direct approaches that use the same photometric loss function. The speed and adaptive nature of our algorithm makes it especially useful in aerial multi-robot applications that can exploit parallel computation.

In this work, we do not investigate robustness against occlusion, leaving it as future work. However, as suggested in~\cite{detone2016}, we could potentially address this issue by using data augmentation techniques such as artificially inserting random occluding shapes into the training images. Another direction for future work is investigating different improvements to achieve sub-pixel accuracy in the top $30\%$ performance percentile. 

Finally, our approach is easily scalable to more general warping motions. Our findings provide additional evidence for applying deep learning methods, specifically unsupervised learning, to various robotic perception problems such as stereo depth estimation, or visual odometry. Our insights on estimating homographies with unsupervised deep neural network approaches provide an initial step in a structured progression of applying these methods to larger problems.

\section{Acknowledgements}
We gratefully acknowledge the support of
ARL grants  W911NF-08-2-0004 
and W911NF-10-2-0016, 
ARO grant W911NF-13-1-0350, 
N00014-14-1-0510, 
N00014-09-1-1051, 
N00014-11-1-0725, 
N00014-15-1-2115 
and N00014-09-1-103, 
DARPA grants HR001151626/HR0011516850 
USDA grant 2015-67021-23857 
 NSF grants IIS-1138847, 
IIS-1426840 
CNS-1446592 
CNS-1521617 
and   IIS-1328805, 
Qualcomm Research,
United Technologies,
and TerraSwarm, one of six centers of STARnet, a Semiconductor Research Corporation program sponsored by MARCO and DARPA. We would also like to thank Aerial Applications for the UAV data set.

%% file: root.bbl
\begin{thebibliography}{10}
\providecommand{\url}[1]{#1}
\csname url@rmstyle\endcsname
\providecommand{\newblock}{\relax}
\providecommand{\bibinfo}[2]{#2}
\providecommand\BIBentrySTDinterwordspacing{\spaceskip=0pt\relax}
\providecommand\BIBentryALTinterwordstretchfactor{4}
\providecommand\BIBentryALTinterwordspacing{\spaceskip=\fontdimen2\font plus
\BIBentryALTinterwordstretchfactor\fontdimen3\font minus
  \fontdimen4\font\relax}
\providecommand\BIBforeignlanguage[2]{{%
\expandafter\ifx\csname l@#1\endcsname\relax
\typeout{** WARNING: IEEEtran.bst: No hyphenation pattern has been}%
\typeout{** loaded for the language `#1'. Using the pattern for}%
\typeout{** the default language instead.}%
\else
\language=\csname l@#1\endcsname
\fi
#2}}

\bibitem{brown2003recognising}
M.~Brown, D.~G. Lowe, \emph{et~al.}, ``Recognising panoramas.'' in \emph{ICCV},
  vol.~3, 2003, p. 1218.

\bibitem{shridhar2015monocular}
M.~Shridhar and K.-Y. Neo, ``Monocular slam for real-time applications on
  mobile platforms,'' 2015.

\bibitem{zhang19963d}
Z.~Zhang and A.~R. Hanson, ``3d reconstruction based on homography mapping,''
  \emph{Proc. ARPA96}, pp. 1007--1012, 1996.

\bibitem{pan2004easy}
Z.~Pan, X.~Fang, J.~Shi, and D.~Xu, ``Easy tour: a new image-based virtual tour
  system,'' in \emph{Proceedings of the 2004 ACM SIGGRAPH international
  conference on Virtual Reality continuum and its applications in
  industry}.\hskip 1em plus 0.5em minus 0.4em\relax ACM, 2004, pp. 467--471.

\bibitem{tang2007self}
C.-Y. Tang, Y.-L. Wu, P.-C. Hu, H.-C. Lin, and W.-C. Chen, ``Self-calibration
  for metric 3d reconstruction using homography.'' in \emph{MVA}, 2007, pp.
  86--89.

\bibitem{capel2004image}
D.~Capel, ``Image mosaicing,'' in \emph{Image Mosaicing and
  Super-resolution}.\hskip 1em plus 0.5em minus 0.4em\relax Springer, 2004, pp.
  47--79.

\bibitem{szeliski2006image}
R.~Szeliski, ``Image alignment and stitching: A tutorial,'' \emph{Foundations
  and Trends{\textregistered} in Computer Graphics and Vision}, vol.~2, no.~1,
  pp. 1--104, 2006.

\bibitem{Lucas:1981:IIR:1623264.1623280}
B.~D. Lucas and T.~Kanade, ``An iterative image registration technique with an
  application to stereo vision,'' in \emph{Proceedings of the 7th International
  Joint Conference on Artificial Intelligence - Volume 2}, ser. IJCAI'81.\hskip
  1em plus 0.5em minus 0.4em\relax San Francisco, CA, USA: Morgan Kaufmann
  Publishers Inc., 1981, pp. 674--679.

\bibitem{baker2004lucas}
S.~Baker and I.~Matthews, ``Lucas-kanade 20 years on: A unifying framework,''
  \emph{International journal of computer vision}, vol.~56, no.~3, pp.
  221--255, 2004.

\bibitem{evangelidis2008parametric}
G.~D. Evangelidis and E.~Z. Psarakis, ``Parametric image alignment using
  enhanced correlation coefficient maximization,'' \emph{IEEE Transactions on
  Pattern Analysis and Machine Intelligence}, vol.~30, no.~10, pp. 1858--1865,
  2008.

\bibitem{yan2014heask}
Q.~Yan, Y.~Xu, X.~Yang, and T.~Nguyen, ``Heask: Robust homography estimation
  based on appearance similarity and keypoint correspondences,'' \emph{Pattern
  Recognition}, vol.~47, no.~1, pp. 368--387, 2014.

\bibitem{lucey2013fourier}
S.~Lucey, R.~Navarathna, A.~B. Ashraf, and S.~Sridharan, ``Fourier lucas-kanade
  algorithm,'' \emph{IEEE transactions on pattern analysis and machine
  intelligence}, vol.~35, no.~6, pp. 1383--1396, 2013.

\bibitem{munoz2015rationalizing}
E.~Mu{\~n}oz, P.~M{\'a}rquez-Neila, and L.~Baumela, ``Rationalizing efficient
  compositional image alignment,'' \emph{International Journal of Computer
  Vision}, vol. 112, no.~3, pp. 354--372, 2015.

\bibitem{lowe2004distinctive}
D.~G. Lowe, ``Distinctive image features from scale-invariant keypoints,''
  \emph{International journal of computer vision}, vol.~60, no.~2, pp. 91--110,
  2004.

\bibitem{Fischler:1981:RSC:358669.358692}
M.~A. Fischler and R.~C. Bolles, ``Random sample consensus: A paradigm for
  model fitting with applications to image analysis and automated
  cartography,'' \emph{Commun. ACM}, vol.~24, no.~6, pp. 381--395, June 1981.

\bibitem{wu2007improved}
F.-l. Wu and X.-y. Fang, ``An improved ransac homography algorithm for feature
  based image mosaic,'' in \emph{Proceedings of the 7th WSEAS International
  Conference on Signal Processing, Computational Geometry \& Artificial
  Vision}.\hskip 1em plus 0.5em minus 0.4em\relax World Scientific and
  Engineering Academy and Society (WSEAS), 2007, pp. 202--207.

\bibitem{rublee2011orb}
E.~Rublee, V.~Rabaud, K.~Konolige, and G.~Bradski, ``Orb: An efficient
  alternative to sift or surf,'' in \emph{Computer Vision (ICCV), 2011 IEEE
  international conference on}.\hskip 1em plus 0.5em minus 0.4em\relax IEEE,
  2011, pp. 2564--2571.

\bibitem{weinzaepfel2013deepflow}
P.~Weinzaepfel, J.~Revaud, Z.~Harchaoui, and C.~Schmid, ``Deepflow: Large
  displacement optical flow with deep matching,'' in \emph{Proceedings of the
  IEEE International Conference on Computer Vision}, 2013, pp. 1385--1392.

\bibitem{ilg2016flownet}
E.~Ilg, N.~Mayer, T.~Saikia, M.~Keuper, A.~Dosovitskiy, and T.~Brox, ``Flownet
  2.0: Evolution of optical flow estimation with deep networks,'' \emph{arXiv
  preprint arXiv:1612.01925}, 2016.

\bibitem{fischer2015flownet}
P.~Fischer, A.~Dosovitskiy, E.~Ilg, P.~H{\"a}usser, C.~Haz{\i}rba{\c{s}},
  V.~Golkov, P.~van~der Smagt, D.~Cremers, and T.~Brox, ``Flownet: Learning
  optical flow with convolutional networks,'' \emph{arXiv preprint
  arXiv:1504.06852}, 2015.

\bibitem{revaud2016deepmatching}
J.~Revaud, P.~Weinzaepfel, Z.~Harchaoui, and C.~Schmid, ``Deepmatching:
  Hierarchical deformable dense matching,'' \emph{International Journal of
  Computer Vision}, vol. 120, no.~3, pp. 300--323, 2016.

\bibitem{altwaijry2016learning}
H.~Altwaijry, A.~Veit, S.~J. Belongie, and C.~Tech, ``Learning to detect and
  match keypoints with deep architectures.'' in \emph{BMVC}, 2016.

\bibitem{eigen2014depth}
D.~Eigen, C.~Puhrsch, and R.~Fergus, ``Depth map prediction from a single image
  using a multi-scale deep network,'' in \emph{Advances in neural information
  processing systems}, 2014, pp. 2366--2374.

\bibitem{detone2016}
D.~DeTone, T.~Malisiewicz, and A.~Rabinovich, ``Deep image homography
  estimation,'' \emph{arXiv preprint arXiv:1606.03798}, 2016.

\bibitem{zhou2017unsupervised}
T.~Zhou, M.~Brown, N.~Snavely, and D.~G. Lowe, ``Unsupervised learning of depth
  and ego-motion from video,'' \emph{arXiv preprint arXiv:1704.07813}, 2017.

\bibitem{baker2006parameterizing}
S.~Baker, A.~Datta, and T.~Kanade, ``Parameterizing homographies,''
  \emph{Technical Report CMU-RI-TR-06-11}, 2006.

\bibitem{simonyan2014very}
K.~Simonyan and A.~Zisserman, ``Very deep convolutional networks for
  large-scale image recognition,'' \emph{arXiv preprint arXiv:1409.1556}, 2014.

\bibitem{zhao2015l2}
H.~Zhao, O.~Gallo, I.~Frosio, and J.~Kautz, ``Is l2 a good loss function for
  neural networks for image processing?'' \emph{ArXiv e-prints}, vol. 1511,
  2015.

\bibitem{iandola2016squeezenet}
F.~N. Iandola, S.~Han, M.~W. Moskewicz, K.~Ashraf, W.~J. Dally, and K.~Keutzer,
  ``Squeezenet: Alexnet-level accuracy with 50x fewer parameters and< 0.5 mb
  model size,'' \emph{arXiv preprint arXiv:1602.07360}, 2016.

\bibitem{Hartley2004}
R.~I. Hartley and A.~Zisserman, \emph{Multiple View Geometry in Computer
  Vision}, 2nd~ed.\hskip 1em plus 0.5em minus 0.4em\relax Cambridge University
  Press, ISBN: 0521540518, 2004.

\bibitem{golub1970singular}
G.~H. Golub and C.~Reinsch, ``Singular value decomposition and least squares
  solutions,'' \emph{Numerische mathematik}, vol.~14, no.~5, pp. 403--420,
  1970.

\bibitem{papadopoulo2000estimating}
T.~Papadopoulo and M.~I. Lourakis, ``Estimating the jacobian of the singular
  value decomposition: Theory and applications,'' in \emph{European Conference
  on Computer Vision}.\hskip 1em plus 0.5em minus 0.4em\relax Springer, 2000,
  pp. 554--570.

\bibitem{hartley2003multiple}
R.~Hartley and A.~Zisserman, \emph{Multiple view geometry in computer
  vision}.\hskip 1em plus 0.5em minus 0.4em\relax Cambridge university press,
  2003.

\bibitem{jaderberg2015spatial}
M.~Jaderberg, K.~Simonyan, A.~Zisserman, \emph{et~al.}, ``Spatial transformer
  networks,'' in \emph{Advances in Neural Information Processing Systems},
  2015, pp. 2017--2025.

\bibitem{abadi2016tensorflow}
M.~Abadi, A.~Agarwal, P.~Barham, E.~Brevdo, Z.~Chen, C.~Citro, G.~S. Corrado,
  A.~Davis, J.~Dean, M.~Devin, \emph{et~al.}, ``Tensorflow: Large-scale machine
  learning on heterogeneous distributed systems,'' \emph{arXiv preprint
  arXiv:1603.04467}, 2016.

\bibitem{kingma2014adam}
D.~Kingma and J.~Ba, ``Adam: A method for stochastic optimization,''
  \emph{arXiv preprint arXiv:1412.6980}, 2014.

\end{thebibliography}
